\documentclass{article}

\usepackage[final]{neurips_2025}

\usepackage[utf8]{inputenc} %
\usepackage[T1]{fontenc}    %
\usepackage{hyperref}       %
\usepackage{url}            %
\usepackage{booktabs}       %
\usepackage{amsfonts}       %
\usepackage{nicefrac}       %
\usepackage{microtype}      %
\usepackage{xcolor}         %
\usepackage{graphicx}       %
\usepackage{svg}
\usepackage{amsmath}
\usepackage{todonotes}
\usepackage{float}
\usepackage{multirow}  %
\usepackage{threeparttable}  %

\newcommand{\x}{\mathbf{x}}
\newcommand{\p}{^{\prime}}

\title{Hash Collisions in Molecular Fingerprints: Effects on Property Prediction and Bayesian Optimization}

\author{%
  Walter Virany \\ %
  University of Toronto\\
  \texttt{wvirany@cs.toronto.edu} \\
  \And
  Austin Tripp \\
  Valence Labs \\
  \texttt{www.austintripp.ca} \\
}

\begin{document}

\maketitle

\begin{abstract}

Molecular fingerprinting methods use hash functions to create fixed-length vector representations of molecules.
However, hash collisions cause distinct substructures to be represented with the same feature,
leading to overestimates in molecular similarity calculations. We investigate whether using exact fingerprints improves accuracy compared to standard compressed fingerprints in molecular property prediction and Bayesian optimization where the underlying predictive model is a Gaussian process. We find that using exact fingerprints yields a small yet consistent improvement in predictive accuracy on five molecular property prediction benchmarks from the DOCKSTRING dataset. However, these gains did not translate to significant improvements in Bayesian optimization performance. 
  
\end{abstract}

\section{Introduction}

Molecular fingerprints are vector representations that encode chemical structure information \citep{yang2022concepts}.
Despite substantial advances in AI, neural network representations of molecules have not been shown
to robustly outperform fingerprints on molecular property prediction tasks,
particularly given limited training data
\citep{green2023current,praski2025benchmarking}.

\begin{figure}[tb]
    \centering
    \includegraphics[width=\linewidth]{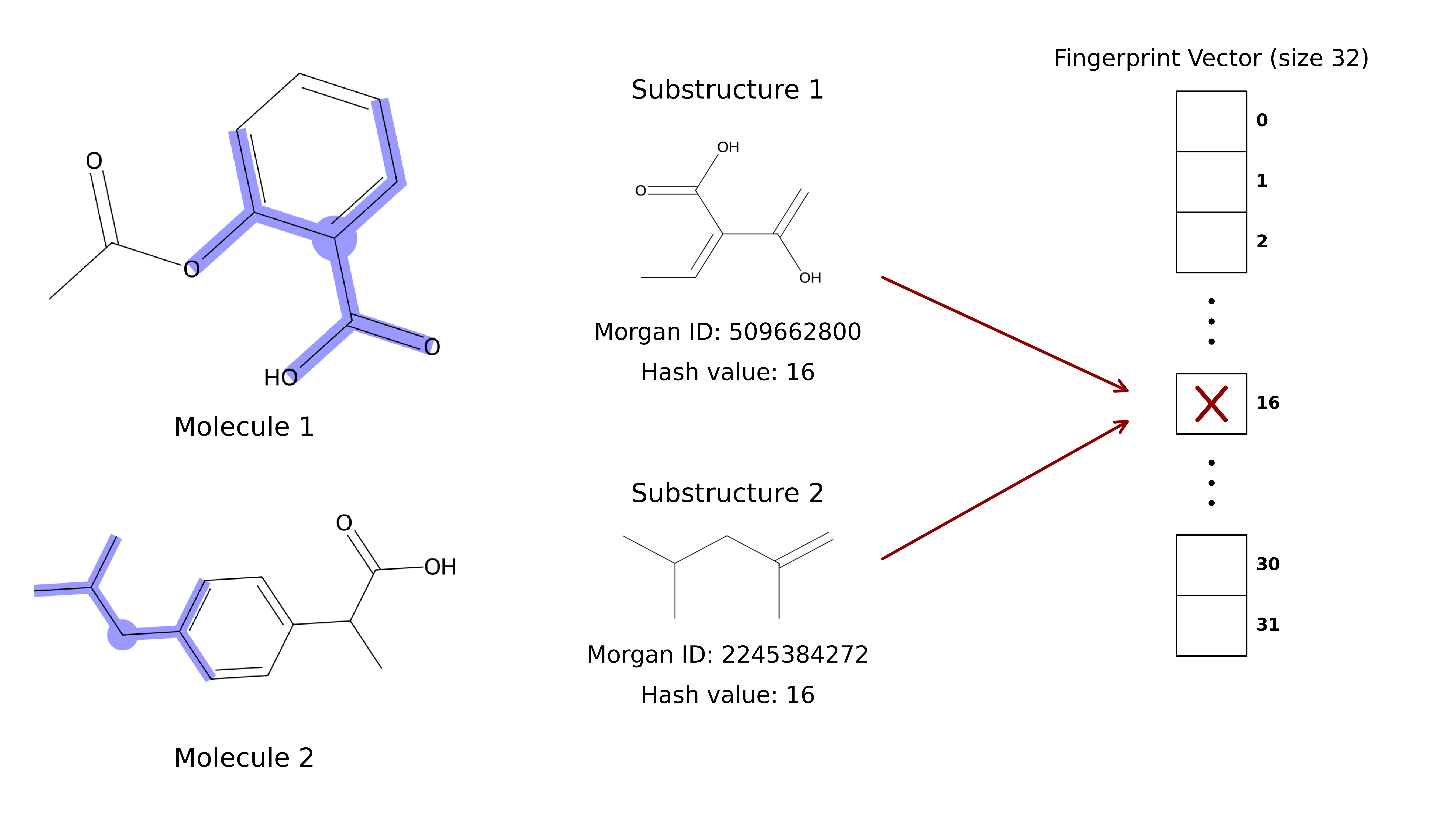}\\
    \caption{Example of collisions. (Left) Two structurally different molecules (SMILES strings are \texttt{CC(=O)OC1=CC=CC=C1C(=O)O} and \texttt{CC(C)CC1=CC=C(C=C1)C(C)C(=O)O}) with highlighted circular substructures of radius 2. (Middle) Highlighted substructures and their corresponding Morgan identifiers, as well as the resulting hash values after taking modulo 32 (corresponding to the fingerprint size). (Right) Fingerprint vector where both distinct substructures map to the same element given by the hash key, demonstrating how different substructures can map to the same dimension.}
    \label{fig:ecfp-collisions}
\end{figure}

Ideally, each dimension in a molecular fingerprint should represent a unique substructure in a molecule.
For example, in 
Extended Connectivity Fingerprints (ECFPs) \citep{rogers2020ecfp},\footnote{
Alternatively known as \emph{Morgan} fingerprints. Arguably, this is the most popular type of fingerprint.
}
each dimension represents a circular substructure up to a given radius.
However, due to the combinatorially large number of possible substructures, such fingerprints would be extremely high dimensional.
Common practice is to instead use a \emph{fixed}-length vector, usually of dimension 1024 or 2048,
using hash functions to map substructure to a dimension.
When there are more structures present than the length of this vector,
it will inevitably result in \emph{collisions}, where a single dimension is used to represent multiple distinct substructures.
An example of this is given in Figure~\ref{fig:ecfp-collisions}.

Intuitively, one would expect the effect of such collisions to be problematic, since it prevents machine learning algorithms from distinguishing between substructures whose effects may be different.
In this paper we investigate this hypothesis quantitatively using Gaussian process (GP) models.
Because GPs are kernel methods which typically use the \emph{similarity} between fingerprints as input,
the effect of hash collisions can be isolated without introducing confounding factors like a changed number of parameters.
Our results in \S\ref{sec:results} show that eliminating collisions results in a small but consistent improvement
on regression metrics, but has a negligible effect on Bayesian optimization (BO) performance.
Given GPs' natural applicability to lab-in-the-loop active learning tasks \citep{moss2020flowmo,griffiths2023gauche},
we believe these results provide important guidance to practitioners.

\section{Background}

\subsection{Gaussian processes}

GPs are a commonly used class of probabilistic models which assume that the data follows a joint multivariate Gaussian distribution.
The mean of the distribution is given by a mean function $\mu: \mathcal{X} \to \mathbb{R}$, and the covariance is given by a positive-definite kernel function $k: \mathcal{X} \times \mathcal{X} \to \mathbb{R}$, where $\mathcal{X}$ denotes some input space.
A detailed treatment of GPs can be found in several textbooks \citep{rasmussen2006gpml, garnett2023bayesopt}.
GPs have become popular for molecular property prediction due to their efficacy in low-data regimes \citep{green2023current, moss2020flowmo, griffiths2023gauche}.
A common choice of kernel function for molecular fingerprints is the \emph{Tanimoto kernel}, which computes the Tanimoto similarity between two molecules, defined as
\begin{equation}
k_{\text{Tanimoto}}(\x, \x\p) = \frac{\sum_i \min\left(x_i, x_i\p\right)} {\sum_i \max\left(x_i, x_i\p\right)} \ ,
\end{equation}
where $\x, \x\p $ are molecular fingerprint vectors. The full covariance function then takes the form
\begin{equation}
k(\x, \x\p) = a^2 \,  k_{\text{Tanimoto}}(\x, \x\p) + \sigma_n^2 \, \delta(\x, \x\p)\ ,
\end{equation}
where $a^2$ and $\sigma_n^2$ are the amplitude and noise variance hyperparameters respectively, and $\delta(\x, \x\p)$ is the Kronecker delta.

\subsection{Bayesian optimization}

BO is a sequential optimization method for expensive black-box functions \citep{garnett2023bayesopt}.
BO uses a probabilistic surrogate model (commonly a GP) to approximate the unknown objective function from observed data.
At each iteration, an acquisition function balances exploration of uncertain regions with exploitation of promising areas to select the next point(s) to evaluate.
In molecular optimization, this corresponds to selecting molecules for expensive experimental evaluation (e.g., synthesis and testing).
The acquired data is used to retrain the surrogate model, yielding a more accurate approximation of the objective function.
This process is then repeated until some stopping criterion is met (e.g., budget exhausted or satisfactory molecules found).

BO heavily relies on accurate predictions of the objective function by the surrogate model.
However, as we established, when using a GP surrogate model, the performance can suffer as a result of overestimates of the similarities between molecules.
This paper investigates whether exact fingerprints improve GP accuracy and whether this has a substantial effect on BO performance.

\section{Methods}

\subsection{Fingerprint representations}

We evaluated three different strategies for encoding molecular fingerprints. Exact fingerprints, as previously mentioned, preserve all unique substructures without compression, resulting in variable-length representations depending on the number of unique substructures present in the molecule. In contrast, compressed fingerprints use fixed-length vectors with RDKit's default hash function, causing collisions when the number of unique substructures exceeds the vector dimension. Finally, we used the Sort\&Slice method as proposed by \citet{dablander2024sortslice}, which selects the most frequently occurring substructures from a reference dataset rather than using hash-based folding. This eliminates collisions for the most common substructures, reducing the number of hash collisions overall.

For all three approaches, we used count-based ECFPs with radius 2. Count fingerprints store the frequency of each substructure, as opposed to binary fingerprints which only indicate the presence/absence of each substructure. Our rationale for this was to include as much information as possible in the compressed fingerprints, in order to not give exact fingerprints an artificial advantage. For the Sort\&Slice method, we calibrated the substructure selection using the ZINC250k dataset. Thus, all three fingerprint methods compute the same underlying substructure identifiers, with differences arising solely from how these are encoded into vectors.

\subsection{Gaussian process implementation}

Standard GP libraries typically require fixed-dimensional input vectors, making them incompatible with exact fingerprints that have variable dimensions. To address this, we developed a custom GP implementation that works directly with RDKit fingerprint objects by computing Tanimoto similarities using the Tanimoto kernel and storing only the resulting kernel matrices. This approach enables fair comparison between exact and fixed-length fingerprints using identical GP methodology with performance differences attributable solely to the fingerprint representation and resulting kernel evaluations.

\subsection{Experiment design}

\textbf{Datasets and tasks.} We evaluated our methods on five protein-ligand docking targets from the DOCKSTRING regression benchmark \citep{garcia-ortegon2022dockstring}. DOCKSTRING provides a dataset of over 260,000 diverse molecules each docked against 58 protein targets, resulting in over 15 million docking scores. The five targets in the regression benchmark span different protein families and difficulty levels: PARP1 (enzyme, easy), F2 (protease, easy-medium), KIT (kinase, medium), ESR2 (nuclear receptor, hard), and PGR (nuclear receptor, hard). We use the predefined cluster-based train/test split to prevent data leakage from structurally similar molecules.

\textbf{Molecular property prediction.} To assess the effects of hash collisions in molecular fingerprints on property prediction, we evaluated GP regression performance on the aforementioned tasks for each fingerprint type. For the fixed-length fingerprint vectors (compressed and Sort\&Slice), we tested the following fingerprint dimensions: 512, 1024, 2048, and 4096. Although the full training set contains approximately 220,000 molecules, we trained on a subset of 10,000 randomly sampled molecules due to the computational expense of GP training, following the same methodology outlined in \citet{garcia-ortegon2022dockstring}. We evaluated predictions using $R^2$, mean squared error (MSE), and mean absolute error (MAE) on the predefined test set consisting of approximately 38,000 molecules. Each experiment was run under two settings of the GP hyperparameters: fixed and optimized (details on hyperparameter settings can be found in Appendix~\ref{app:hyperparam_optimization}). Results were averaged across 10 random trials for each setting.

\textbf{Bayesian optimization.} We also investigated the effect of hash collisions in BO. Our hypothesis was that even small differences in predictive accuracy would compound, yielding significant differences in BO outcomes over many iterations. To test this, we simulated BO on the same five DOCKSTRING targets using a GP surrogate model with fixed hyperparameters. We used the complete dataset (combining training and test sets) as the candidate pool, randomly selected 1,000 molecules from the bottom 80\% of the dataset as the initial training set, and ran optimization with a budget of 1,000 iterations using expected improvement (EI) as the acquisition function. The same three fingerprint configurations previously discussed were compared, using fingerprint dimensions of 1024 and 2048 for the fixed-length fingerprints. We evaluated performance using area under the curve (AUC) of the best observed value over time, and averaged our results across 5 random trials of each experiment setting.

\section{Results}\label{sec:results}

\subsection{Molecular property prediction}

Exact fingerprints consistently outperformed compressed fingerprints across all DOCKSTRING targets under both hyperparameter settings. Table~\ref{tab:regression_summary} shows results with optimized hyperparameters, where exact fingerprints universally outperformed the \emph{best}-scoring compressed configurations, with improvements in $R^2$ scores ranging from 0.006 (F2) to 0.017 (KIT). Against Sort\&Slice fingerprints, exact fingerprints showed mixed performance: outperforming on two targets (ESR2, KIT), matching performance on two targets (F2, PARP1), and being outperformed on one target (PGR), where Sort\&Slice achieved an $R^2$ score of 0.480 vs. 0.470. Notably, PGR was both the hardest task (with the lowest overall $R^2$ scores) and the only target where Sort\&Slice outperformed exact fingerprints, achieving its best performance at the smallest dimension tested (512) rather than the typical 4096. These same patterns were observed across all evaluation metrics (MSE and MAE) and were particularly pronounced under fixed hyperparameter settings, where exact fingerprints outperformed all alternatives on every target. Complete results are provided in Appendix~\ref{app:full_results}.

\begin{table}[b]
\centering
\caption{GP Regression Performance with Optimized Hyperparameters.
Values show mean $R^2$ scores $\pm1$ standard deviation across 10 random train/test splits. For compressed and Sort\&Slice fingerprints, numbers in parentheses indicate the best-performing fingerprint dimension among those tested (512, 1024, 2048, and 4096). Bold indicates best performance for each target. An asterisk $^\ast$ indicates statistically significant difference from exact fingerprint based on Tukey's HSD test ($\alpha = 0.05)$.
}
\label{tab:regression_summary}
\begin{tabular}{lccc}
\toprule
Target & Exact & Compressed & Sort\&Slice \\
\midrule
ESR2   & $\textbf{0.553} \pm 0.003$ & $^\ast0.54 \pm 0.003$ \scriptsize (4096) & $^\ast 0.545 \pm 0.004$ \scriptsize (4096) \\
F2     & $\textbf{0.848} \pm 0.002$ & $^\ast 0.842 \pm 0.002$ \scriptsize (4096) & $\textbf{0.848} \pm 0.002$ \scriptsize (4096) \\
KIT    & $\textbf{0.745} \pm 0.002$ & $^\ast0.728 \pm 0.002$ \scriptsize (4096) & $^\ast0.739 \pm 0.003$ \scriptsize (4096) \\
PARP1  & $\textbf{0.882} \pm 0.001$ & $^\ast0.876 \pm 0.001$ \scriptsize (4096) & $\textbf{0.882} \pm 0.001$ \scriptsize (4096) \\
PGR    & $0.470 \pm 0.004$ & $^\ast0.457 \pm 0.004$ \scriptsize (4096) & $^\ast\textbf{0.480} \pm 0.003$ \scriptsize (512) \\
\bottomrule
\end{tabular}
\end{table}

To assess statistical significance, we performed one-way ANOVA followed by Tukey's HSD post hoc test with type I error rate $\alpha = 0.05$, comparing exact fingerprints against each alternative. Exact fingerprints showed statistically significant performance improvements against compressed fingerprints across all targets, fingerprint sizes, and hyperparameter settings. For Sort\&Slice, the performance differences observed were statistically significant on ESR2, KIT, and PGR, while differences on F2 and PARP1 were not significant.

\begin{figure}[t]
    \centering
    \includegraphics[width=\linewidth]{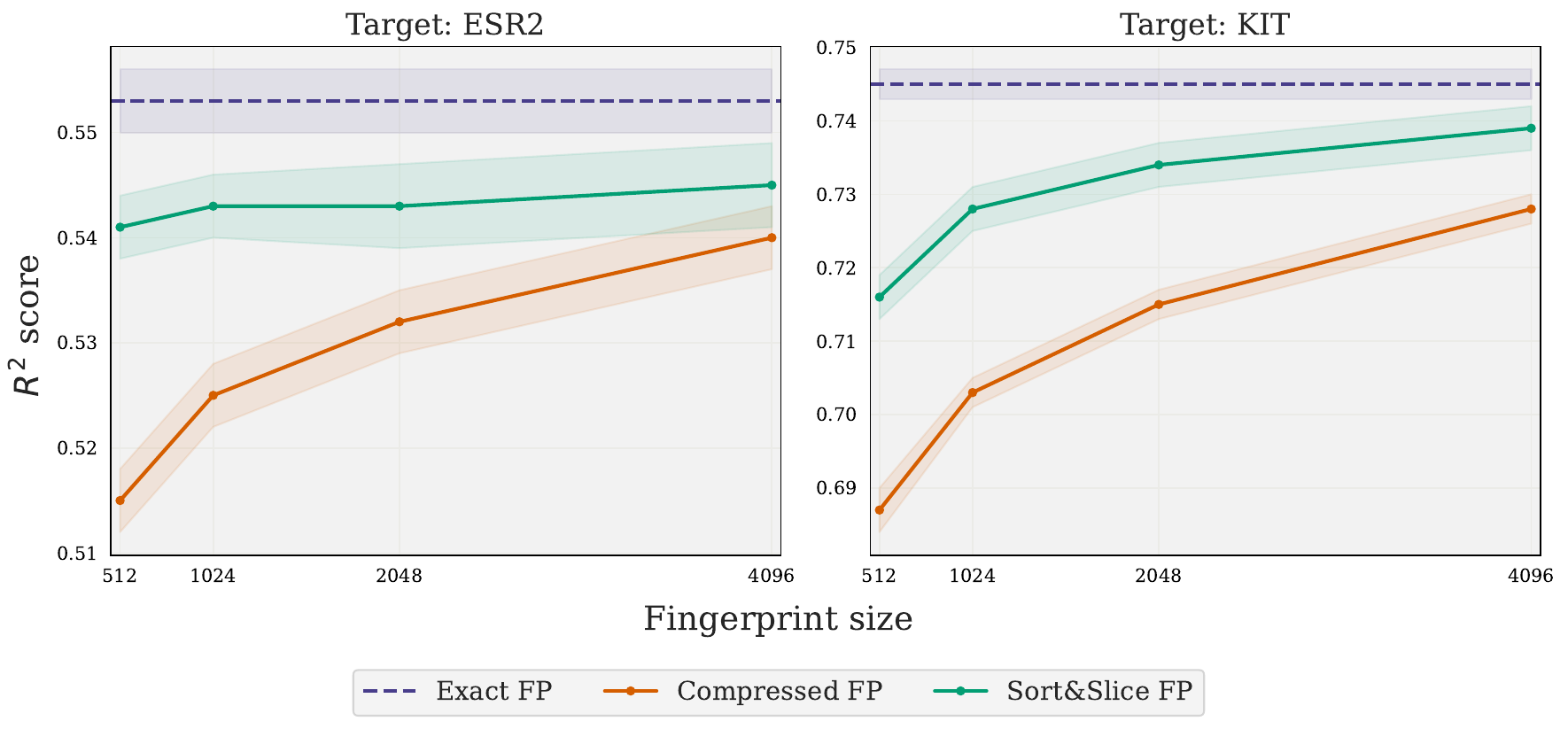}\\
    \caption{$R^2$ scores for GP regression with optimized hyperparameters as a function of fingerprint dimension on ESR2 (left) and KIT (right) targets. Exact fingerprints (purple) consistently outperform compressed fingerprints (orange) and Sort\&Slice method (green). Dark lines indicate mean and shaded regions indicate $\pm1$ standard deviation across 10 random train/test splits. Note that the dimension of exact fingerprints is not changing, but the performance is included as a horizontal line for reference.}
    \label{fig:regression}
\end{figure}

Figure~\ref{fig:regression} demonstrates the monotonic relationship between fingerprint dimension and predictive performance for the ESR2 and KIT targets. Fixed-length fingerprints show consistent improvements in $R^2$ score as dimension increases from 512 to 4096, approaching exact fingerprint performance. This scaling pattern was observed for both compressed and Sort\&Slice fingerprints across nearly all targets and hyperparameter settings, with the sole exception of Sort\&Slice on PGR with optimized hyperparameters. These observations are consistent with the hypothesis that mitigating errors in similarity calculations leads to improved regression performance in GPs. We provide further analysis of hash collision rates and the resulting effects on similarity calculations in Appendix~\ref{app:hash_collision_rates}.

\subsection{Bayesian optimization}

Despite the consistent improvements in GP regression accuracy, exact fingerprints yielded no significant improvement in BO performance. Table~\ref{tab:bo_summary} summarizes AUC scores across all five targets, showing no clear advantage for any fingerprint configuration. In fact, exact fingerprints achieved the best performance on only one target (ESR2), while compressed and Sort\&Slice methods matched or exceeded exact performance on the remaining targets.

\begin{table}[b]
\centering
\caption{
Bayesian Optimization Performance (AUC scores).
Values show mean $\pm1$ standard deviation across 5 random trials. Bold indicates best performance for each target.
}
\label{tab:bo_summary}
\begin{tabular}{lcccccc}
\toprule
& & \multicolumn{2}{c}{Compressed} & \multicolumn{2}{c}{Sort\&Slice} \\
\cmidrule(lr){3-4} \cmidrule(lr){5-6}
Target & Exact & 1024 & 2048 & 1024 & 2048 \\
\midrule
ESR2   & $\textbf{0.973} \pm 0.004$ & $0.965 \pm 0.015$ & $0.952 \pm 0.013$ & $0.968 \pm 0.01$3 & $0.972 \pm 0.006$ \\
F2     & $0.986 \pm 0.004$ & $\textbf{0.989} \pm 0.004$ & $\textbf{0.989} \pm 0.004$ & $0.983 \pm 0.005$ & $0.988 \pm 0.004$ \\
KIT    & $0.973 \pm 0.007$ & $0.973 \pm 0.01$ & $0.971 \pm 0.005$ & $0.978 \pm 0.005$ & $\textbf{0.981} \pm 0.005$ \\
PARP1  & $0.972 \pm 0.004$ & $0.970 \pm 0.003$ & $0.977 \pm 0.002$ & $0.977 \pm 0.002$ & $\textbf{0.978} \pm 0.006$ \\
PGR    & $0.860 \pm 0.023$ & $0.868 \pm 0.035$ & $0.856 \pm 0.028$ & $0.867 \pm 0.021$ & $\textbf{0.873} \pm 0.02$3 \\
\bottomrule
\end{tabular}
\end{table}

\begin{figure}
    \centering
    \includegraphics[width=\linewidth]{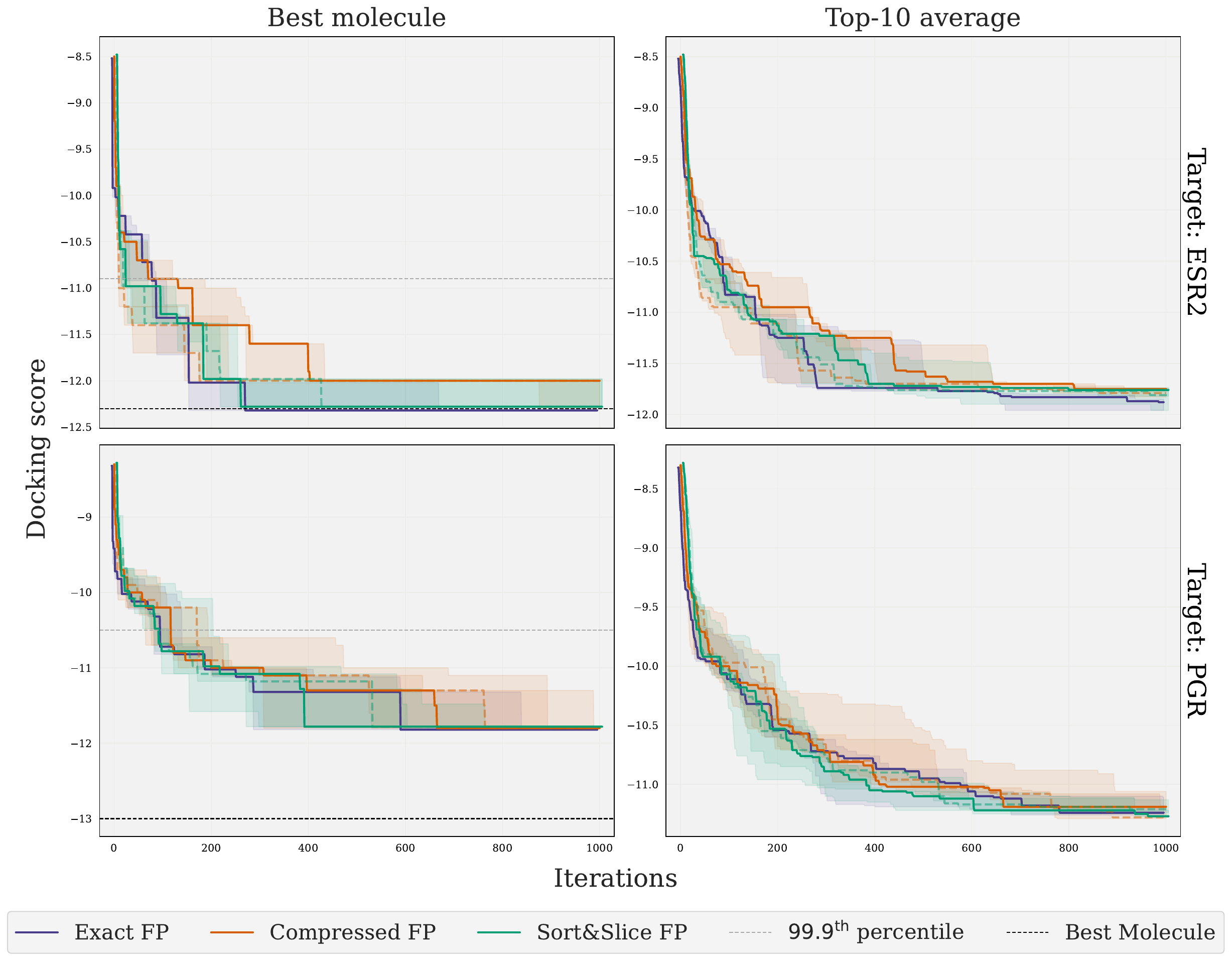}\\
    \caption{BO trajectories for two targets: ESR2 (top) and PGR (bottom). The first column shows the score of the best molecule at each iteration, and the second column shows the average score of the top 10 acquired molecules at each iteration. For fixed-length fingerprint types, two fingerprint sizes are shown: 1024 (dashed line) and 2048 (solid line). Dark lines indicate the median and shaded regions indicate the 1st and 3rd quartiles over 5 random trials. The dashed horizontal lines in the left-hand figures indicate the 99.9$^{\text{th}}$ percentile and the best possible score for each target.} 
    \label{fig:bo1}
\end{figure}

Figure~\ref{fig:bo1} shows representative BO trajectories for two DOCKSTRING targets, illustrating how all fingerprint configurations follow similar optimization paths despite their different regression accuracies. This disconnect suggests that the small improvements observed in predictive accuracy do not necessarily translate to significant differences in the sequential decision-making process that drives molecular optimization.

\section{Discussion}

Our results demonstrate that exact fingerprints consistently improve GP regression performance across all DOCKSTRING targets, supporting our hypothesis that hash collisions lead to overestimated molecular similarities and thus reduced predictive accuracy. The Sort\&Slice method performs comparably to exact fingerprints, suggesting that preserving the most informative substructures may be an effective alternative to eliminating all hash collisions. Interestingly, these regression improvements did not translate to enhanced Bayesian optimization performance, where all fingerprint types achieved similar AUC scores. Future work could further investigate the relationship between BO and predictive accuracy of the surrogate model. We note that, in addition to predictive performance, accurate uncertainty estimates are also a key component of successful BO procedures.

Several limitations should be acknowledged. The Sort\&Slice method performance may be overestimated due to potential data leakage, as we calibrated this method using the ZINC250k dataset, which was also the dataset used for training and evaluation. Additionally, the optimization landscapes may not have been sufficiently challenging to reveal differences in BO between fingerprint types. For example, for even the hardest target (PGR), all BO procedures found top 0.1\% molecules within several hundred iterations. In addition to addressing these, future work could investigate the effects of hash collisions in other types of fingerprints or different kernel methods such as SVMs. While preliminary results did show similar observations with different fingerprint types, a systematic analysis might prove valuable. Despite these limitations, our work suggests that for property prediction tasks, methods which reduce hash collisions in molecular fingerprints provide modest but consistent improvements, while their benefits for BO remain unclear.

\bibliography{references}
\bibliographystyle{plainnat}

\appendix

\clearpage
\section{Hash collisions analysis}\label{app:hash_collision_rates}

To quantify the prevalence of hash collisions in molecular fingerprints, we analyzed the frequency of collisions in 10,000 pairs of molecules randomly sampled from the DOCKSTRING dataset, and the resulting effect on Tanimoto similarity calculations.

\begin{table}[h]
\centering
\caption{Pairwise collision statistics and Tanimoto similarity differences for compressed fingerprints. Values are reported as the mean over 10,000 randomly sampled pairs of molecules. Overestimation is measured as the difference between compressed and exact Tanimoto similarity.}
\label{tab:pairwise_collisions}
\begin{tabular}{lcccc}
\toprule
\textbf{} & \textbf{512} & \textbf{1024} & \textbf{2048} & \textbf{4096} \\
\midrule
Pairwise collisions & $4.87$ & $2.43$ & $1.41$ & $0.48$ \\
\midrule
Exact Tanimoto similarity & $0.167$ & $0.167$ & $0.167$ & $0.167$ \\
Compressed Tanimoto similarity & $0.198$ & $0.183$ & $0.176$ & $0.171$ \\
\midrule
Overestimation & $0.031$ & $0.0152$ & $0.008$ & $0.004$ \\
\bottomrule
\end{tabular}
\end{table}

\begin{figure}
    \centering
    \includegraphics[width=\linewidth]{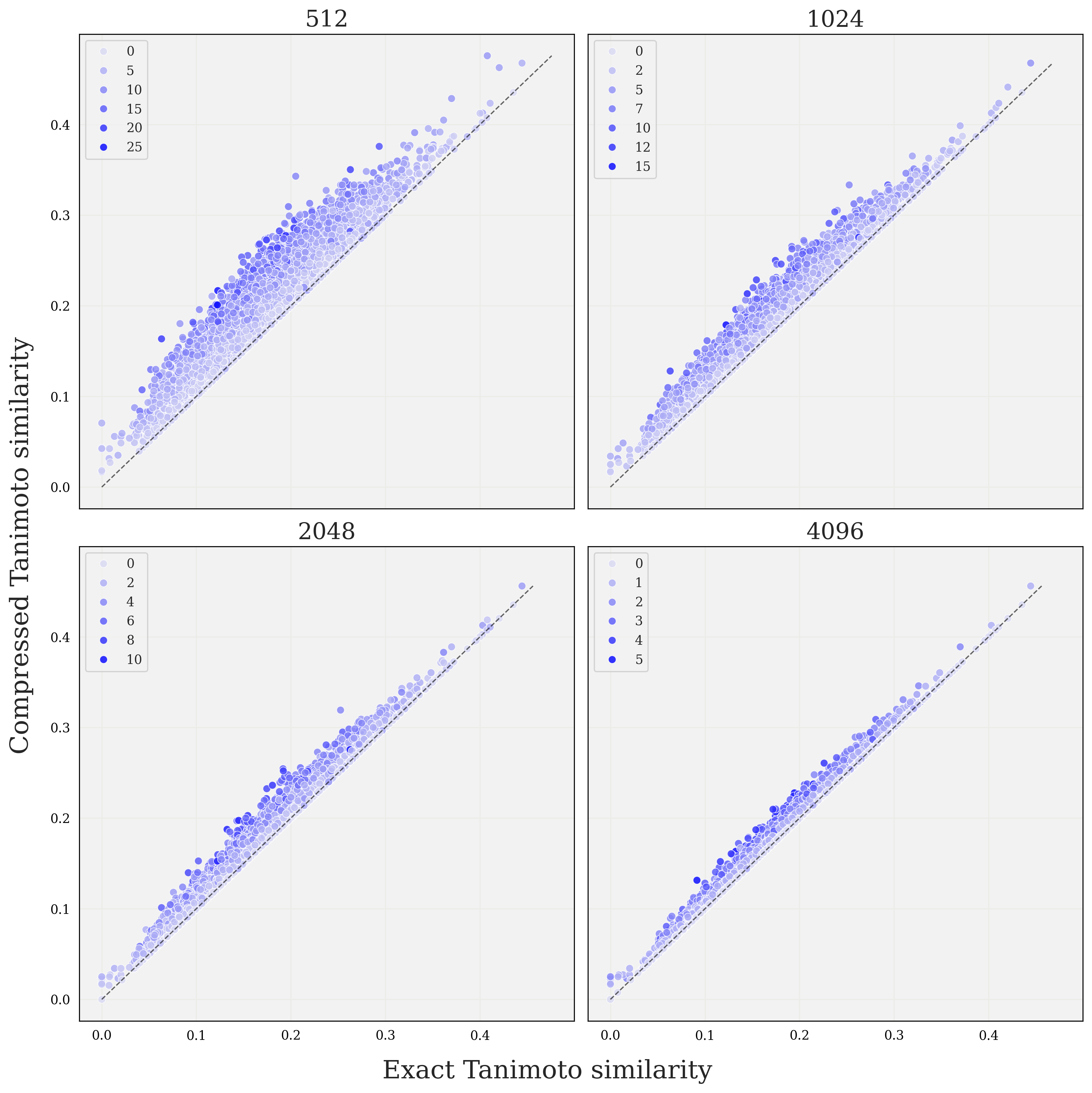}\\
    \caption{Tanimoto similarity in compressed vs. exact fingerprints for 10,000 pairs of molecules. Each point represents one pair, where the x-axis is the Tanimoto similarity computed between exact fingerprints, and the y-axis is the Tanimoto similarity computed between compressed fingerprints. The diagonal line corresponds to equal similarity calculations for the two fingerprints, and points above this line indicate that compression results in overestimated similarity. Shade indicates the number of hash collisions between each pair, with darker colors representing more collisions.} 
    \label{fig:collisions}
\end{figure}

Table~\ref{tab:pairwise_collisions} demonstrates the inverse relationship between fingerprint size and the number of hash collisions between pairs of molecules. 
As the fingerprint size increases, the number of hash collisions decreases, along with the overestimation in Tanimoto similarity. Figure ~\ref{fig:collisions} visualizes this systematic overestimation, where molecule pairs with more collisions tend to lie further above the diagonal line corresponding to equal similarity calculations.

\section{GP hyperparameter selection}\label{app:hyperparam_optimization}

We evaluated GP performance under two hyperparameter settings: fixed and optimized hyperparameters.

\textbf{Fixed hyperparameters.} For the fixed hyperparameter setting, we used the following values across all experiments:

\begin{itemize}
    \item Amplitude: Set to the empirical variance of the training targets for each dataset
    \item Noise variance: Set to $0.01 \times$ amplitude
    \item Mean function: Constant mean set to the empirical mean of the training targets 
\end{itemize}

These values represent commonly used heuristics in GP regression and ensure consistent comparison across different fingerprint representations.

\textbf{Optimized hyperparameters.} For the optimized setting, we learned the hyperparameters by optimizing the marginal log-likelihood under gradient descent. Our optimization procedure is outlined as follows:

\begin{enumerate}
    \item Initialization: Started with the same values used for fixed hyperparameters described above
    \item Optimizer: Used Adam optimizer with learning rate 0.01
    \item Objective: Maximized the marginal log-likelihood of the training data
    \item Convergence criteria: Gradient norm below $10^{-3}$ or maximum 10,000 iterations reached
    \item Numerical stability: Applied lower bounds on noise variance (minimum $10^{-4}\,\times$ target variance) to prevent numerical instability
\end{enumerate}

The optimization was performed independently for each combination of fingerprint type, dataset, and random trial, allowing each configuration to find its optimal hyperparameters. This approach ensures that performance differences between fingerprint types are not confounded by suboptimal hyperparameter choices.

For the BO experiments, we used only the fixed hyperparameter setting to maintain computational efficiency during the sequential optimization process.

\section{Complete regression results}\label{app:full_results}

This section provides comprehensive regression results across all DOCKSTRING targets, fingerprint dimensions, and hyperparameter settings. All values represent mean $\pm$ 1 standard deviation across 10 random trials. Bold indicates the best performance for each target.

\subsection{Optimized hyperparameters}

\begin{table}[H]
\centering
\caption{$R^2$ scores with optimized hyperparameters ($\uparrow$)}
\tiny
\setlength{\tabcolsep}{3pt}
\begin{tabular}{l|c|cccc|cccc}
\toprule
\multirow{2}{*}{Target} & \multirow{2}{*}{Exact} & \multicolumn{4}{c|}{Compressed} & \multicolumn{4}{c}{Sort\&Slice} \\[2pt]
& & 512 & 1024 & 2048 & 4096 & 512 & 1024 & 2048 & 4096 \\
\midrule
ESR2 & \textbf{0.553 $\pm$ 0.003} & 0.515 $\pm$ 0.003& 0.525 $\pm$ 0.003& 0.532 $\pm$ 0.003& 0.540 $\pm$ 0.003 & 0.541 $\pm$ 0.003& 0.543 $\pm$ 0.003& 0.543 $\pm$ 0.004& 0.545 $\pm$ 0.004 \\[3pt]
F2 & \textbf{0.848 $\pm$ 0.002} & 0.83 $\pm$ 0.002& 0.834 $\pm$ 0.001& 0.838 $\pm$ 0.002& 0.842 $\pm$ 0.002 & 0.843 $\pm$ 0.002& 0.845 $\pm$ 0.001& 0.846 $\pm$ 0.001& \textbf{0.848 $\pm$ 0.002} \\[3pt]
KIT & \textbf{0.745 $\pm$ 0.002} & 0.687 $\pm$ 0.003& 0.703 $\pm$ 0.002& 0.715 $\pm$ 0.002& 0.728 $\pm$ 0.002 & 0.716 $\pm$ 0.003& 0.728 $\pm$ 0.003& 0.734 $\pm$ 0.003& 0.739 $\pm$ 0.003 \\[3pt]
PARP1 & \textbf{0.882 $\pm$ 0.001} & 0.865 $\pm$ 0.001& 0.869 $\pm$ 0.001& 0.872 $\pm$ 0.001& 0.876 $\pm$ 0.001 & 0.878 $\pm$ 0.001& 0.879 $\pm$ 0.001& 0.881 $\pm$ 0.001& \textbf{0.882 $\pm$ 0.001} \\[3pt]
PGR & 0.470 $\pm$ 0.004 & 0.444 $\pm$ 0.004& 0.444 $\pm$ 0.004& 0.449 $\pm$ 0.004& 0.457 $\pm$ 0.004 & \textbf{0.480 $\pm$ 0.003} & 0.478 $\pm$ 0.002& 0.476 $\pm$ 0.003& 0.474 $\pm$ 0.003\\
\bottomrule
\end{tabular}
\label{tab:opt_r2}
\end{table}

\begin{table}[H]
\centering
\caption{MSE with optimized hyperparameters ($\downarrow$)}
\tiny
\setlength{\tabcolsep}{3pt}
\begin{tabular}{l|c|cccc|cccc}
\toprule
\multirow{2}{*}{Target} & \multirow{2}{*}{Exact} & \multicolumn{4}{c|}{Compressed} & \multicolumn{4}{c}{Sort\&Slice} \\[2pt]
& & 512 & 1024 & 2048 & 4096 & 512 & 1024 & 2048 & 4096 \\
\midrule
ESR2 & \textbf{0.348 $\pm$ 0.002}& 0.378 $\pm$ 0.002& 0.37 $\pm$ 0.002& 0.364 $\pm$ 0.002& 0.358 $\pm$ 0.002& 0.357 $\pm$ 0.002& 0.356 $\pm$ 0.003& 0.355 $\pm$ 0.003& 0.354 $\pm$ 0.003\\[3pt]
F2 & \textbf{0.156 $\pm$ 0.002}& 0.176 $\pm$ 0.002& 0.171 $\pm$ 0.002& 0.167 $\pm$ 0.002& 0.163 $\pm$ 0.002& 0.162 $\pm$ 0.002& 0.16 $\pm$ 0.001& 0.159 $\pm$ 0.001& 0.157 $\pm$ 0.002\\[3pt]
KIT & \textbf{0.3 $\pm$ 0.003}& 0.369 $\pm$ 0.003& 0.351 $\pm$ 0.002& 0.336 $\pm$ 0.002& 0.321 $\pm$ 0.003& 0.335 $\pm$ 0.003& 0.321 $\pm$ 0.003& 0.314 $\pm$ 0.003& 0.307 $\pm$ 0.004\\[3pt]
PARP1 & \textbf{0.185 $\pm$ 0.001}& 0.213 $\pm$ 0.001& 0.206 $\pm$ 0.001& 0.201 $\pm$ 0.001& 0.195 $\pm$ 0.001& 0.192 $\pm$ 0.001& 0.19 $\pm$ 0.001& 0.187 $\pm$ 0.001& 0.186 $\pm$ 0.001\\[3pt]
PGR & 0.494 $\pm$ 0.003& 0.519 $\pm$ 0.004& 0.519 $\pm$ 0.003& 0.514 $\pm$ 0.003& 0.506 $\pm$ 0.004& \textbf{0.485 $\pm$ 0.003}& 0.487 $\pm$ 0.002& 0.489 $\pm$ 0.003& 0.49 $\pm$ 0.003\\
\bottomrule
\end{tabular}
\label{tab:opt_mse}
\end{table}

\begin{table}[H]
\centering
\caption{MAE with optimized hyperparameters ($\downarrow$)}
\tiny
\setlength{\tabcolsep}{3pt}
\begin{tabular}{l|c|cccc|cccc}
\toprule
\multirow{2}{*}{Target} & \multirow{2}{*}{Exact} & \multicolumn{4}{c|}{Compressed} & \multicolumn{4}{c}{Sort\&Slice} \\[2pt]
& & 512 & 1024 & 2048 & 4096 & 512 & 1024 & 2048 & 4096 \\
\midrule
ESR2 & \textbf{0.449 $\pm$ 0.002}& 0.469 $\pm$ 0.001& 0.465 $\pm$ 0.001& 0.461 $\pm$ 0.001& 0.457 $\pm$ 0.002& 0.456 $\pm$ 0.002& 0.455 $\pm$ 0.002& 0.455 $\pm$ 0.002& 0.454 $\pm$ 0.002\\[3pt]
F2 & \textbf{0.305 $\pm$ 0.002}& 0.325 $\pm$ 0.002& 0.32 $\pm$ 0.001& 0.316 $\pm$ 0.002& 0.311 $\pm$ 0.002& 0.311 $\pm$ 0.002& 0.309 $\pm$ 0.001& 0.308 $\pm$ 0.002& 0.306 $\pm$ 0.002\\[3pt]
KIT & \textbf{0.422 $\pm$ 0.002}& 0.47 $\pm$ 0.002& 0.458 $\pm$ 0.001& 0.448 $\pm$ 0.001& 0.437 $\pm$ 0.002& 0.446 $\pm$ 0.002& 0.437 $\pm$ 0.002& 0.432 $\pm$ 0.002& 0.427 $\pm$ 0.002\\[3pt]
PARP1 & \textbf{0.326 $\pm$ 0.001}& 0.352 $\pm$ 0.001& 0.346 $\pm$ 0.001& 0.341 $\pm$ 0.001& 0.335 $\pm$ 0.001& 0.332 $\pm$ 0.001& 0.33 $\pm$ 0.001& 0.328 $\pm$ 0.001& 0.327 $\pm$ 0.001\\[3pt]
PGR & 0.553 $\pm$ 0.002& 0.568 $\pm$ 0.002& 0.568 $\pm$ 0.002& 0.565 $\pm$ 0.002& 0.561 $\pm$ 0.002& \textbf{0.545 $\pm$ 0.002}& 0.547 $\pm$ 0.002& 0.549 $\pm$ 0.003& 0.55 $\pm$ 0.002\\
\bottomrule
\end{tabular}
\label{tab:opt_mae}
\end{table}

\subsection{Fixed hyperparameters}

\begin{table}[H]
\centering
\caption{$R^2$ scores with fixed hyperparameters ($\uparrow$)}
\tiny
\setlength{\tabcolsep}{3pt}
\begin{tabular}{l|c|cccc|cccc}
\toprule
\multirow{2}{*}{Target} & \multirow{2}{*}{Exact} & \multicolumn{4}{c|}{Compressed} & \multicolumn{4}{c}{Sort\&Slice} \\[2pt]
& & 512 & 1024 & 2048 & 4096 & 512 & 1024 & 2048 & 4096 \\
\midrule
ESR2 & \textbf{0.533 $\pm$ 0.004}& 0.491 $\pm$ 0.003& 0.498 $\pm$ 0.004& 0.5 $\pm$ 0.003& 0.507 $\pm$ 0.004& 0.49 $\pm$ 0.004& 0.501 $\pm$ 0.005& 0.506 $\pm$ 0.005& 0.511 $\pm$ 0.005\\[3pt]
F2 & \textbf{0.832 $\pm$ 0.003}& 0.815 $\pm$ 0.003& 0.818 $\pm$ 0.003& 0.821 $\pm$ 0.003& 0.824 $\pm$ 0.003& 0.823 $\pm$ 0.003& 0.826 $\pm$ 0.002& 0.827 $\pm$ 0.002& 0.828 $\pm$ 0.003\\[3pt]
KIT & \textbf{0.732 $\pm$ 0.003}& 0.675 $\pm$ 0.003& 0.688 $\pm$ 0.002& 0.699 $\pm$ 0.002& 0.712 $\pm$ 0.002& 0.695 $\pm$ 0.003& 0.71 $\pm$ 0.003& 0.719 $\pm$ 0.003& 0.727 $\pm$ 0.003\\[3pt]
PARP1 & \textbf{0.867 $\pm$ 0.002}& 0.852 $\pm$ 0.001& 0.855 $\pm$ 0.001& 0.857 $\pm$ 0.001& 0.86 $\pm$ 0.002& 0.863 $\pm$ 0.001& 0.865 $\pm$ 0.002& 0.866 $\pm$ 0.001& 0.866 $\pm$ 0.002\\[3pt]
PGR & \textbf{0.461 $\pm$ 0.005}& 0.435 $\pm$ 0.005& 0.434 $\pm$ 0.005& 0.432 $\pm$ 0.005& 0.439 $\pm$ 0.006& 0.452 $\pm$ 0.005& 0.456 $\pm$ 0.004& 0.458 $\pm$ 0.004& 0.459 $\pm$ 0.005\\
\bottomrule
\end{tabular}
\label{tab:fixed_r2}
\end{table}

\begin{table}[H]
\centering
\caption{MSE with fixed hyperparameters ($\downarrow$)}
\tiny
\setlength{\tabcolsep}{3pt}
\begin{tabular}{l|c|cccc|cccc}
\toprule
\multirow{2}{*}{Target} & \multirow{2}{*}{Exact} & \multicolumn{4}{c|}{Compressed} & \multicolumn{4}{c}{Sort\&Slice} \\[2pt]
& & 512 & 1024 & 2048 & 4096 & 512 & 1024 & 2048 & 4096 \\
\midrule
ESR2 & \textbf{0.363 $\pm$ 0.003}& 0.397 $\pm$ 0.003& 0.391 $\pm$ 0.003& 0.389 $\pm$ 0.003& 0.384 $\pm$ 0.003& 0.397 $\pm$ 0.003& 0.389 $\pm$ 0.004& 0.385 $\pm$ 0.004& 0.381 $\pm$ 0.004\\[3pt]
F2 & \textbf{0.174 $\pm$ 0.003}& 0.191 $\pm$ 0.003& 0.188 $\pm$ 0.003& 0.185 $\pm$ 0.003& 0.182 $\pm$ 0.003& 0.182 $\pm$ 0.003& 0.18 $\pm$ 0.003& 0.179 $\pm$ 0.003& 0.177 $\pm$ 0.003\\[3pt]
KIT & \textbf{0.315 $\pm$ 0.003}& 0.383 $\pm$ 0.004& 0.367 $\pm$ 0.002& 0.354 $\pm$ 0.002& 0.339 $\pm$ 0.003& 0.36 $\pm$ 0.004& 0.342 $\pm$ 0.004& 0.331 $\pm$ 0.004& 0.322 $\pm$ 0.004\\[3pt]
PARP1 & \textbf{0.21 $\pm$ 0.003}& 0.233 $\pm$ 0.002& 0.228 $\pm$ 0.002& 0.225 $\pm$ 0.002& 0.22 $\pm$ 0.003& 0.215 $\pm$ 0.002& 0.213 $\pm$ 0.002& 0.211 $\pm$ 0.002& \textbf{0.21 $\pm$ 0.002}\\[3pt]
PGR & \textbf{0.503 $\pm$ 0.004}& 0.527 $\pm$ 0.005& 0.528 $\pm$ 0.005& 0.529 $\pm$ 0.005& 0.523 $\pm$ 0.005& 0.511 $\pm$ 0.005& 0.507 $\pm$ 0.004& 0.505 $\pm$ 0.004& 0.504 $\pm$ 0.004\\
\bottomrule
\end{tabular}
\label{tab:fixed_mse}
\end{table}

\begin{table}[H]
\centering
\caption{MAE with fixed hyperparameters ($\downarrow$)}
\tiny
\setlength{\tabcolsep}{3pt}
\begin{tabular}{l|c|cccc|cccc}
\toprule
\multirow{2}{*}{Target} & \multirow{2}{*}{Exact} & \multicolumn{4}{c|}{Compressed} & \multicolumn{4}{c}{Sort\&Slice} \\[2pt]
& & 512 & 1024 & 2048 & 4096 & 512 & 1024 & 2048 & 4096 \\
\midrule
ESR2 & \textbf{0.46 $\pm$ 0.002}& 0.481 $\pm$ 0.002& 0.478 $\pm$ 0.002& 0.478 $\pm$ 0.002& 0.474 $\pm$ 0.002& 0.482 $\pm$ 0.002& 0.477 $\pm$ 0.003& 0.475 $\pm$ 0.003& 0.472 $\pm$ 0.003\\[3pt]
F2 & \textbf{0.319 $\pm$ 0.002}& 0.336 $\pm$ 0.002& 0.333 $\pm$ 0.002& 0.33 $\pm$ 0.002& 0.327 $\pm$ 0.003& 0.329 $\pm$ 0.002& 0.327 $\pm$ 0.002& 0.325 $\pm$ 0.002& 0.324 $\pm$ 0.002\\[3pt]
KIT & \textbf{0.432 $\pm$ 0.002}& 0.478 $\pm$ 0.002& 0.468 $\pm$ 0.002& 0.46 $\pm$ 0.001& 0.449 $\pm$ 0.002& 0.464 $\pm$ 0.002& 0.451 $\pm$ 0.002& 0.444 $\pm$ 0.002& 0.438 $\pm$ 0.003\\[3pt]
PARP1 & \textbf{0.342 $\pm$ 0.002}& 0.366 $\pm$ 0.001& 0.36 $\pm$ 0.001& 0.357 $\pm$ 0.001& 0.352 $\pm$ 0.002& 0.351 $\pm$ 0.001& 0.349 $\pm$ 0.001& 0.347 $\pm$ 0.001& 0.346 $\pm$ 0.001\\[3pt]
PGR & 0.556 $\pm$ 0.003& 0.569 $\pm$ 0.003& 0.57 $\pm$ 0.003& 0.57 $\pm$ 0.003& 0.567 $\pm$ 0.003& 0.556 $\pm$ 0.003& \textbf{0.554 $\pm$ 0.002} & \textbf{0.554 $\pm$ 0.002}& \textbf{0.554 $\pm$ 0.003}\\
\bottomrule
\end{tabular}
\label{tab:fixed_mae}
\end{table}

\section{Code and reproducibility}

The custom GP implementation can be found at \href{https://github.com/AustinT/tanimoto-gp}{https://github.com/AustinT/tanimoto-gp}. All experiments from the paper can be found at \href{https://github.com/wvirany/molcollisions}{https://github.com/wvirany/molcollisions}.

\end{document}